%% file: main.tex
\documentclass[letterpaper]{article} % DO NOT CHANGE THIS
\usepackage{aaai24}  % DO NOT CHANGE THIS
\usepackage{times}  % DO NOT CHANGE THIS
\usepackage{helvet}  % DO NOT CHANGE THIS
\usepackage{courier}  % DO NOT CHANGE THIS
\usepackage[hyphens]{url}  % DO NOT CHANGE THIS
\usepackage{graphicx} % DO NOT CHANGE THIS
\usepackage{natbib}  % DO NOT CHANGE THIS AND DO NOT ADD ANY OPTIONS TO IT
\usepackage{caption} % DO NOT CHANGE THIS AND DO NOT ADD ANY OPTIONS TO IT
\usepackage{algorithm}
\usepackage{algorithmic}

\usepackage{newfloat}
\usepackage{listings}
\DeclareCaptionStyle{ruled}{labelfont=normalfont,labelsep=colon,strut=off} % DO NOT CHANGE THIS
\floatstyle{ruled}
\newfloat{listing}{tb}{lst}{}
\floatname{listing}{Listing}
\def\alambicdeit{\includegraphics[width=0.025\linewidth]{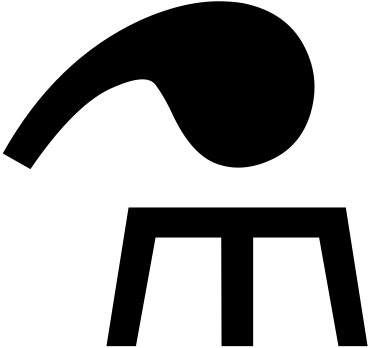}}

\usepackage{times}
\usepackage{epsfig}
\usepackage{graphicx}
\usepackage{amsmath}
\usepackage{amssymb}
\newcommand*{\myalign}[2]{\multicolumn{1}{#1}{#2}}

\usepackage[utf8]{inputenc} % allow utf-8 input
\usepackage[T1]{fontenc}    % use 8-bit T1 fonts
\usepackage{url}            % simple URL typesetting
\usepackage{booktabs}       % professional-quality tables
\usepackage{amsfonts}       % blackboard math symbols
\usepackage{nicefrac}       % compact symbols for 1/2, etc.
\usepackage{microtype}      % microtypography

\usepackage{afterpage}
\usepackage{amsmath}
\usepackage{amsthm}
\usepackage{csquotes}
\usepackage{enumitem}
\usepackage{graphicx}
\usepackage{lipsum}
\usepackage{booktabs}
\usepackage[labelformat=simple]{subcaption}
\usepackage{array}

\usepackage{pgfgantt}
\usepackage{multirow}
\usepackage{array,booktabs}
\usepackage{tabularray}
\usepackage{soul}

\usepackage{svg}
\usepackage{pgfplots}
\usepackage{multirow, tabularx}
\usepackage{mathtools}
\usepackage{cancel}
\usepackage{adjustbox}
\usepackage{xcolor}

\usepackage{epsfig}
\usepackage{graphicx}
\usepackage{amsmath}
\usepackage{amssymb}
\usepackage{colortbl}
\usepackage{stackengine}
\usepackage{listings}

\definecolor{light_red}{rgb}{0.921,0.78039,0.83137}
\definecolor{dark_red}{rgb}{0.729,0.459,0.459}

\newcommand*{\belowrulesepcolor}[1]{% 
  \noalign{% 
    \kern-\belowrulesep 
    \begingroup 
      \color{#1}% 
      \hrule height\belowrulesep 
    \endgroup 
  }%
} 
\newcommand*{\aboverulesepcolor}[1]{% 
  \noalign{% 
    \begingroup 
      \color{#1}% 
      \hrule height\aboverulesep 
    \endgroup 
    \kern-\aboverulesep 
  }%
} 

\newcommand{\lpnorm}[1]{\left\lVert #1 \right\rVert}

\newcommand{\Real}{{\rm I\!R}}

\definecolor{mygrey}{RGB}{230,230,230} 
\definecolor{myblue}{RGB}{230,230,255}
\definecolor{codegreen}{rgb}{0,0.5,0}
\definecolor{codegray}{rgb}{0.5,0.5,0.5}
\definecolor{codepurple}{rgb}{0.58,0,0.82}
\definecolor{backcolour}{rgb}{1.0,1.0,1.0}
\definecolor{link}{rgb}{1.0,1.0,1.0}
\lstdefinestyle{mystyle}{
    backgroundcolor=\color{backcolour},   
    commentstyle=\color{codegreen},
    keywordstyle=\color{magenta},
    numberstyle=\tiny\color{codegray},
    stringstyle=\color{codepurple},
    basicstyle=\ttfamily\footnotesize,
    breakatwhitespace=false,         
    breaklines=true,                 
    keepspaces=true,                 
    numbers=left,                    
    numbersep=5pt,                  
    showspaces=false,                
    showstringspaces=false,
    showtabs=false,                  
    tabsize=1,
    float=tp,
  floatplacement=tbp
}
\DeclareCaptionFormat{mylst}{\hrule#1#2#3\hrule}
\definecolor{light_orange}{RGB}{252,219,191}
\definecolor{dark_orange}{RGB}{237,139,55}
\definecolor{light_green}{RGB}{111,193,174}
\definecolor{dark_green}{RGB}{67,154,134}
\definecolor{cyan}{RGB}{123,250,241}

\definecolor{xarch}{rgb}{1.0, 0.9, 0.9}
\definecolor{sarch}{rgb}{0.94, 0.97, 1.0}
\definecolor{light_green}{rgb}{0.72, 0.85, 0.85}
\definecolor{light_red}{rgb}{1.0, 0.9, 0.9}

\usepackage{tikz}
\usepackage{tabularx}
\usetikzlibrary{fit}

\usepackage{amssymb}% http://ctan.org/pkg/amssymb
\usepackage{pifont}% http://ctan.org/pkg/pifont

\newcounter{nodecount}
\newcommand\tabnode[1]{\addtocounter{nodecount}{1} \tikz \node  (\arabic{nodecount}) {#1};}

\tikzstyle{every picture}+=[remember picture,baseline]
\tikzstyle{every node}+=[anchor=base,minimum width=0.4cm,align=center,text depth=.25ex,outer sep=1.5pt]
\tikzstyle{every path}+=[thick, rounded corners]

\title{Understanding the Role of the Projector in Knowledge Distillation}
\author {
    Roy Miles,
    Krystian Mikolajczyk
}
\affiliations {
   Imperial College London \\
    r.miles18@imperial.ac.uk, k.mikolajczyk@imperial.ac.uk
}

\begin{document}

\maketitle

\begin{abstract}
In this paper we revisit the efficacy of knowledge distillation as a function matching and metric learning problem. In doing so we verify three important design decisions, namely the normalisation, soft maximum function, and projection layers as key ingredients. We theoretically show that the projector implicitly encodes information on past examples, enabling relational gradients for the student. We then show that the normalisation of representations is tightly coupled with the training dynamics of this projector, which can have a large impact on the students performance. Finally, we show that a simple soft maximum function can be used to address any significant capacity gap problems. Experimental results on various benchmark datasets demonstrate that using these insights can lead to superior or comparable performance to state-of-the-art knowledge distillation techniques, despite being much more computationally efficient. In particular, we obtain these results across image classification (CIFAR100 and ImageNet), object detection (COCO2017), and on more difficult distillation objectives, such as training data efficient transformers, whereby we attain a $77.2$\% top-1 accuracy with DeiT-Ti on ImageNet. Code and models are publicly available.
\end{abstract}

\section{Introduction}
\label{sec:intro}
Deep neural networks have achieved remarkable success in various applications, ranging from computer vision~\cite{Krizhevsky2012ImageNetNetworks} to natural language processing~\cite{Vaswani2017AttentionNeed}. However, the high computational cost and memory requirements of deep models have limited their deployment in resource-constrained environments. Knowledge distillation is a popular technique to address this problem through transferring the knowledge of a large teacher model to that of a smaller student model. This technique involves training the student to imitate the output of the teacher, either by directly minimizing the difference between intermediate features or by minimizing the Kullback-Leibler (KL) divergence between their soft predictions. Although knowledge distillation has shown to be very effective, there are still some limitations related to the computational and memory overheads in constructing and evaluating the losses, as well as an insufficient theoretical explanation for the underlying core principles.

\begin{figure}[ht]
    \centering
    \resizebox{1.\columnwidth}{!}{%
    \includegraphics[width=1.\linewidth]{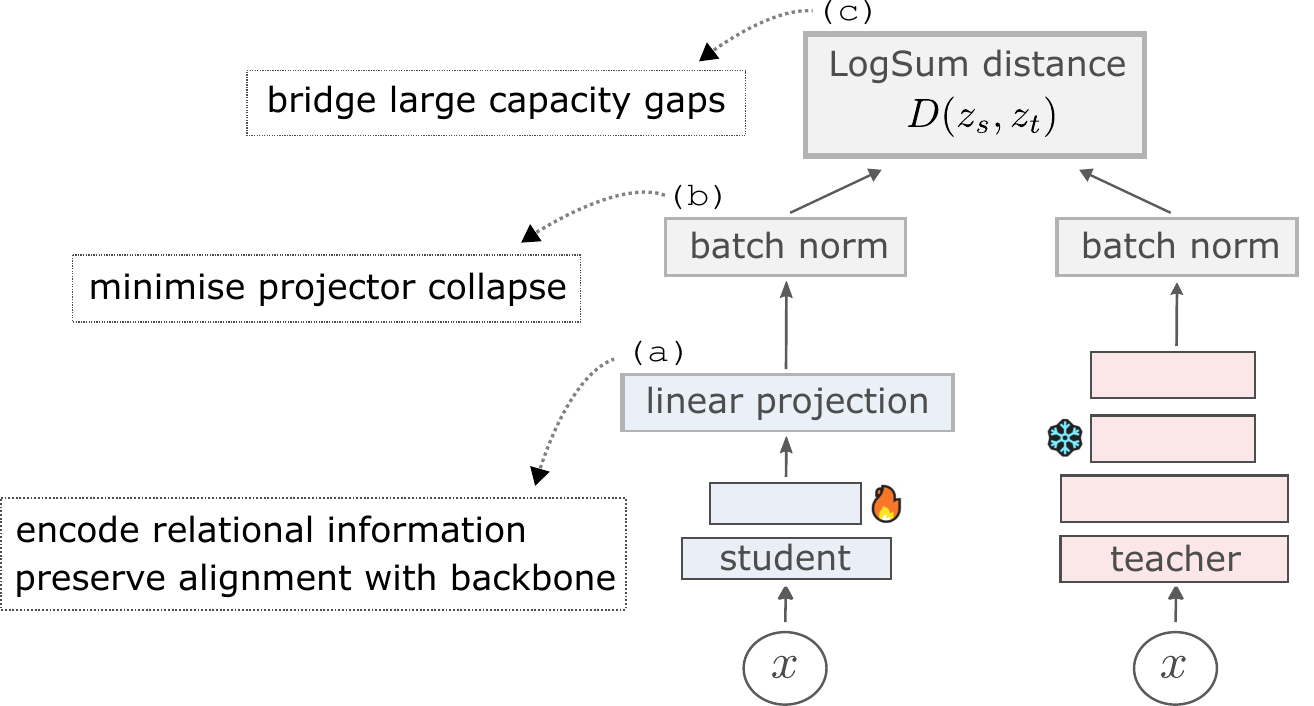}
    }
    \caption{Proposed feature distillation pipeline using three distinct components: linear projection (a), batch norm (b), and a \textit{LogSum} distance (c). We provide an interpretable explanation for each each of these three components, which results in a very cheap and effective recipe for distillation.}
    \vspace{-1em}
    \label{fig:distillation_overview}
\end{figure}

To overcome these limitations, we revisit knowledge distillation from both a function matching and metric learning perspective. We perform an extensive ablation of three important components of knowledge distillation, namely the distance metric, normalisation, and projector network. Alongside this ablation we provide a theoretical perspective and unification of these design principles through exploring the underlying training dynamics. Finally, we extend these principles to a few large scale vision tasks, whereby we achieve comparable or improved performance over state-of-the-art. The most significant result of which pertains to the data-efficient training of transformers, whereby a performance gain of 2.2\% is achieved over the best-performing distillation methods that are designed explicitly for this task.
Our main contributions can be summarised as follows.
\begin{itemize}
    \item We explore three distinct design principles from knowledge distillation, namely the projection, normalisation, and distance function. In doing so we demonstrate their and coupling with each other, both through analytical means and by observing the training dynamics.
    \item We show that a projection layer implicitly encodes relational information from previous samples. Using this knowledge we can remove the need to explicitly construct correlation matrices or memory banks that will inevitably incur a significant memory overhead.
    \item We propose a simple recipe for knowledge distillation using a linear projection, batch normalisation, and a $LogSum$ function. These three design choices can attain competitive or improved performance to state-of-the-art for image classification, object detection, and the data efficient training of transformers. 
\end{itemize}

%We expect that these id and should in fact be more commonly adopted than standard KL divergence practice.
% \begin{itemize}
%     \item This paper highlights that distillation is a form of function matching and demonstrates how a simple metric can be used to transfer knowledge between different tasks.
%     \item We introduce a matching loss between the predictions of interpolated hidden states to improve the robustness and convergence of knowledge distillation.
%     \item The experimental results demonstrate the superiority of the proposed distillation pipeline approach over state-of-the-art knowledge distillation techniques across a range of tasks and distillation settings.
% \end{itemize}

\section{Related Work}
\label{sec:related_work}

\paragraph{Knowledge Distillation}
Knowledge distillation is the process of transferring the knowledge from a large, complex model to a smaller, simpler model. Its usage was originally proposed in the context of image classification~\cite{Hinton2015DistillingNetwork} whereby the soft teacher predictions would encode relational information between classes. Spherical KD~\cite{Guo2020ReducingDistillation} extended this idea by re-scaling the logits, prime aware adaptive distillation~\cite{Zhang2020Prime-AwareDistillation} introduced an adaptive weighting strategy, while DKD~\cite{Zhao2022DecoupledDistillation} proposed to decouple the original formulation into target class and non-target class probabilities.

Hinted losses~\cite{Romero2015FitNets:Nets} were a natural extension of the logit-based approach whereby the intermediate feature maps are used as hints for the student. Attention transfer~\cite{Zagoruyko2019PayingTransfer} then proposed to re-weight this loss using spatial attention maps. ReviewKD~\cite{Chen2021DistillingReview} addressed the problem relating to the arbitrary selection of layers by aggregating information across all the layers using trainable attention blocks. Neuron selectivity transfer~\cite{Huang2017LikeTransfer}, similarity-preserving
KD~\cite{Tung2019Similarity-preservingDistillation}, and relational KD~\cite{Park2019RelationalDistillation} construct relational batch and feature matrices that can be used as inputs for the distillation losses. Similarly FSP matrices~\cite{Yim2017ALearning} were proposed to extract the relational information through a residual block. In contrast to this theme, we show that a simple projection layer can implicitly capture most relational information, thus removing the need to construct any expensive relational structures.

Representation distillation was originally proposed alongside a contrastive based loss~\cite{Tian2019ContrastiveDistillation} and has since been extended using a Wasserstein distance~\cite{Chen2020WassersteinDistillation}, information theory~\cite{Miles2022InformationDistillation}, graph theory~\cite{Ma2022DistillingAlignment}, and complementary gradient information~\cite{Zhu2021ComplementaryDistillation}. Distillation also been empirically shown to benefit from longer training schedules and more data-augmentation~\cite{Beyer2022KnowledgeConsistent}, which is similarly observed with HSAKD~\cite{Yang2021HierarchicalDistillation} and SSKD~\cite{Xu2020KnowledgeSelf-supervision}.
Distillation between CNNs and transformers has also been a very practically motivated task for data-efficient training~\cite{Touvron2021TrainingAttention} and has shown to benefit from an ensemble of teacher architectures~\cite{Ren2022Co-advise:Distillation}. However, we show that just a simple extension of some fundamental distillation design principles is much more effective.

Self-distillation is another branch of knowledge distillation that instead proposes to instead distill knowledge within the network itself~\cite{Zhang2019BeDistillation}. This paradigm removes the need to have any pre-trained teacher readily available and has since been applied in the context of deep metric learning~\cite{roth2021s2sd}, graph neural networks~\cite{chen2021selfdistilling}, and image classification~\cite{Miles2020CascadedSelf-distillation}. Its success of which has since driven many theoretical advancements~\cite{mobahi2020selfdistillation, allen-zhu2023towards, zhang2020selfdistillation} and is still an active area of research.

\paragraph{Self-Supervised Learning}
Self-supervised learning (SSL) is an increasingly popular field of machine learning whereby a model is trained to learn a useful representation of unlabelled data. Its popularity has been driven by the increasing cost of manual labelling and has since been crucial for training large transformer models. Various pretext tasks have been proposed to learn these representations, such as image inpainting~\cite{He2022MaskedLearners}, colorization~\cite{Zhang2016ColorfulColorization}, or prediction of the rotation~\cite{Gidaris2018UnsupervisedRotations} or position of patches~\cite{Doersch2015UnsupervisedPrediction, Carlucci2019DomainPuzzles}. SimCLR~\cite{Chen2020ARepresentations} approached self-supervision using a contrastive loss with multi-view augmentation to define the positive and negative pairs. They found a large memory bank of negative representations was necessary to achieve good performance, but would incur a significant memory overhead. MoCo~\cite{He2020MomentumLearning} extending this work with a momentum encoder, which was subsequently extended by MoCov2~\cite{Chen2020ImprovedLearning} and MoCov3~\cite{Chen2021AnTransformers}.
% that would then introduce various incremental design changes to improve the performance.

% Asymmetric architectures were proposed as an alternative to contrastive learning, whereby no negative samples are needed. Most noticeable works in this area are BYOL~\cite{Grill2020BootstrapLearning} and SimSiam~\cite{Chen2021ExploringLearning} which both use stop gradients to avoid representation collapse. DirectPred~\cite{Tian2021UnderstandingPairs} provided a theoretical understanding of this non-contrastive SSL setting. They introduced a series of conceptual insights into the functional roles of many crucial ingredients used. In doing so they derived a simple linear predictor with competitive performance to some much more complex predictor architectures. In our work we explore knowledge distillation from a similar perspective as DirectPred but observe some unique observations and results pertaining to this distillation setting.

Feature decorrelation is another approach to SSL that avoids the need for negative pairs to address representation collapse. Both Barlow twins~\cite{Zbontar2021BarlowReduction} and VICReg~\cite{Bardes2022VICReg:Learning} achieve this by maximising the variance within a batch, while preserving invariance to augmentations. Both contrastive learning and feature decorrelation have since been extended to dense prediction tasks~\cite{Bardes2022VICRegL:Features} and unified with knowledge distillation~\cite{Miles2023MobileVOS:Distillation}.

\begin{figure*}
\centering
\begin{minipage}[b]{.32\textwidth}
\centering
\includegraphics[width=1.\textwidth]{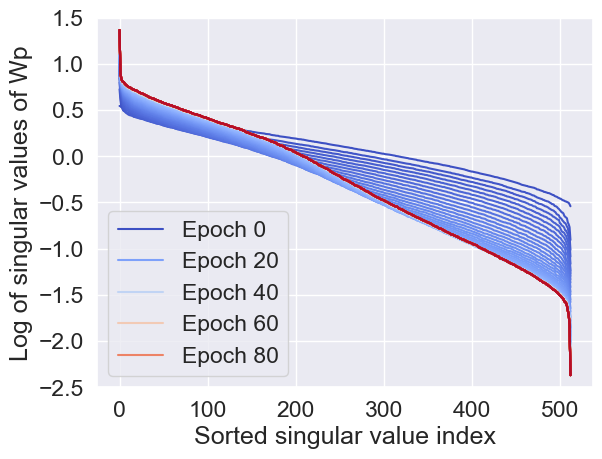}
\caption*{(a) no normalisation.}
\end{minipage}
\hfill
\begin{minipage}[b]{.32\textwidth}
\centering
\includegraphics[width=1.\textwidth]{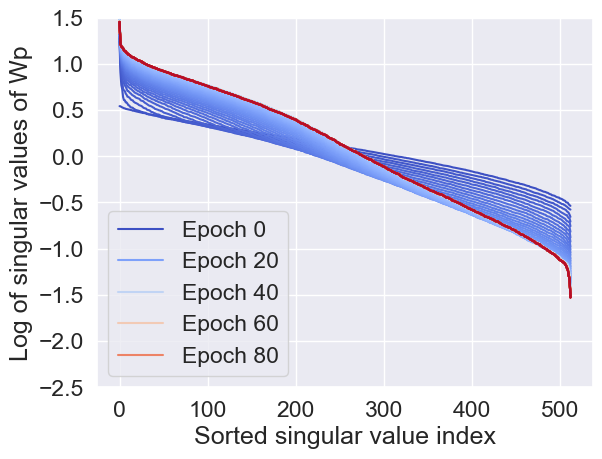}
\caption*{(b) L2 normalisation.}
\end{minipage}
\hfill
\begin{minipage}[b]{.32\textwidth}
\centering
\includegraphics[width=1.\textwidth]{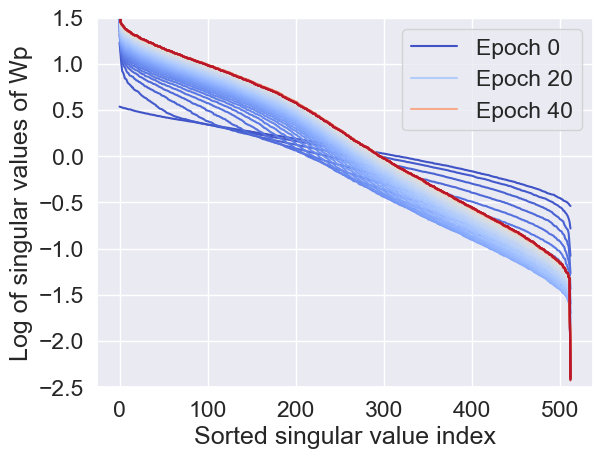}
\caption*{(c) batch normalisation.}
\end{minipage}
\caption{Evolution of singular values of the projection weights $\mathbf{W}_p$ under three different representation normalisation schemes. The student is a Resnet-18, while the teacher is a ResNet-50. The three curves shows the evolution of singular values for the projector weights when the representations undergo no normalisation, L2 normalisation, and batch norm respectively.}
\vspace{-0.4em}
\label{fig:singular_values_evolution}
\end{figure*}

% \section{Understanding the Training Dynamics}
\section{Understanding the Role of the Projector}
\label{sec:understanding_kd}

Knowledge distillation (KD) is a technique used to transfer knowledge from a large, powerful model (teacher) to a smaller, less powerful one (student). In the classification setting, it can be done using the soft teacher predictions as pseudo-labels for the student. Unfortunately, this approach does not trivially generalise to non-classification tasks~\cite{Liu2019StructuredSegmentation} and the classifier may collapse a lot of information~\cite{Tishby2015DeepPrinciple} that can be useful for distillation.
Another approach is to use feature maps from the earlier layers for distillation~\cite{Romero2015FitNets:Nets, Zagoruyko2019PayingTransfer}, however, its usage presents two primary challenges: the difficulty in ensuring consistency across different architectures~\cite{Chen2021DistillingReview} and the potential degradation in the student's downstream performance for cases where the inductive biases of the two networks differ~\cite{Tian2019ContrastiveDistillation}
A compromise, which strikes a balance between the two approaches  discussed above, is to distill the representation directly before the output space. Representation distillation has been successfully adopted in past works~\cite{Tian2019ContrastiveDistillation, Zhu2021ComplementaryDistillation, Miles2022InformationDistillation} and is the focus of this paper. The exact training framework used is described in figure \ref{fig:distillation_overview}. The projection layer shown was originally used to simply match the student and teacher dimensions~\cite{Romero2015FitNets:Nets}, however, we will show that its role is much more important and it can lead to significant performance improvements even when the two feature dimensions already match. The two representations are typically both followed by some normalisation scheme as a way of appropriately scaling the gradients. However, we find this normalisation has a more interesting property in its relation to what information is encoded in the learned projector weights.

In this work we provide a theoretical perspective to motivate some simple and effective design choices for knowledge distillation. In contrast to the recent works~\cite{Tung2019Similarity-preservingDistillation, Miles2022InformationDistillation}, we show that an explicit construction of complex relational structures, such as feature kernels~\cite{He2022FeatureDistillation} is not necessary. In fact, most of this structure can be learned implicitly through the tightly coupled interaction of a learnable projection layer and an appropriate normalisation scheme.
In the following sections we investigate the training dynamics of the projection layer with the choice of normalisation scheme.  We explore the impact and trade-offs that arise from the architecture design of the projector. Finally, we propose a simple modification to the distance metric to address issues arising from a large capacity gap between the student and teacher models. Although we do not aim to necessarily propose a new method for distillation, we uncover a cheap and simple recipe that can transfer to various distillation settings and tasks. Furthermore, we provide a new theoretical perspective on the underlying principles of distillation that can translate to large scale vision tasks. \\

\noindent\textbf{The projection weights encode relational information from previous samples.} The projection layer plays a crucial role in KD as it provides an implicit encoding of previous samples and its weights can capture the relational information needed to transfer information regarding the correlation between features. We observe that even a single linear projector layer can provide significant improvements in accuracy (see Supplementary). This improvement suggests that the projections role in distillation can be described more concisely as being an encoder of essential information needed for the distillation loss itself. Most recent works propose a manual construction of some relational information to be used as part of a loss~\cite{Park2019RelationalDistillation, Tung2019Similarity-preservingDistillation}, however, we posit that an implicit and learnable approach is much more effective. To explore this phenomenon in more detail, we consider the update equations for the projector weights and its training dynamics. Without loss in generality, consider a simple L2 loss and a linear bias-free projection layer.

\begin{align}
    D(\mathbf{Z}_s, \mathbf{Z}_t \;; \mathbf{W}_p) &= \frac{1}{2} \lpnorm{ \mathbf{Z}_s\mathbf{W}_p - \mathbf{Z}_t }_2^2
 \end{align}
Where $\mathbf{Z}_s$ and $\mathbf{Z}_t$ are the \textbf{s}tudent and \textbf{t}eacher representations respectively, while $\mathbf{W}_p$ is the matrix representing the linear \textbf{p}rojection. Using the trace property of the Frobenius norm, we can then express this loss as follows:
\begin{align}
    D(\mathbf{Z}_s, \mathbf{Z}_t \;; \mathbf{W}_p) &= \frac{1}{2}tr\left( \left(\mathbf{Z}_s\mathbf{W}_p - \mathbf{Z}_t\right)^T\left(\mathbf{Z}_s\mathbf{W}_p - \mathbf{Z}_t\right) \right) \\
    &= \frac{1}{2}tr( \mathbf{W}_p^T\mathbf{Z}_s^T\mathbf{Z}_s\mathbf{W}_p - \mathbf{Z}_t^T\mathbf{Z}_s\mathbf{W}_p \\ &\;\;\;\;\;\;\;\;\;\;\;\;-\mathbf{W}_p^T\mathbf{Z}_s^t\mathbf{Z}_t + \mathbf{Z}_t^T\mathbf{Z}_t)
    \label{eqn:dloss}
\end{align}
Taking the derivative with respect to $\mathbf{W}_p$, we can derive the update rule $\dot{\mathbf{W}_p}$
\begin{align}
    \dot{\mathbf{W}_p} &= -\frac{\partial D(\mathbf{W}_p)}{\partial \mathbf{W}_p} = -\mathbf{Z}_s^T\mathbf{Z}_s\mathbf{W}_p + \mathbf{Z}_s^T\mathbf{Z}_t
\end{align}
which can be further simplified
\begin{equation}
    \boxed{\dot{\mathbf{W}_p} = \mathbf{C}_{st} - \mathbf{C}_s\mathbf{W}_p}
    \label{eqn:projector_update}
\end{equation}
where $\mathbf{C}_s = \mathbf{Z}_s^T\mathbf{Z}_s \in \Real^{d_s \times d_s}$ and $\mathbf{C}_{st} = \mathbf{Z}_s^T\mathbf{Z}_t \in \Real^{d_s \times d_t}$ denote self and cross correlation matrices respectively. Due to the capacity gap between the student network and the teacher network, there is no perfect linear projection between these two representation spaces. Instead, the projector will converge on an approximate projection that we later show is governed by the normalisation being employed. 

\underline{Whitened features:} consider using self-supervised learning in conjunction with distillation whereby the student features are whitened to have perfect decorrelation~\cite{Ermolov2020WhiteningLearning}, or alternatively, they are batch normalised and sufficiently regularised with a feature decorrelation term~\cite{Bardes2022VICReg:Learning}. In this setting, the fixed point solution for the weights will be symmetric and will capture the cross relationship between student and teacher features.
\vspace{-0.9em}

\begin{align}
    \mathbf{C}_{st} &- \mathbf{C}_s\mathbf{W}_p = 0  \;\;\; \text{where} \;\;\; \mathbf{C}_s = \mathbf{I} \\
    &\rightarrow \mathbf{W}_p = \mathbf{C}_{st}
\end{align}

Other normalisation schemes, such as those that jointly normalise the projected features and the teacher features, will have a much more involved analysis but will unlikely provide any additional insights on the dynamics of training itself. Thus, we propose to empirically explore the training trajectories of the projector weights singular values. This exploration will help quantify how the projector is mapping the student features to the teachers space. We cover this in the next section along with some additional insights into what is being learned and distilled.

\textbf{The choice of normalisation directly affects the training dynamics of $\mathbf{W}_p$.}
Equation \ref{eqn:projector_update} shows that the projector weights can encode relational information between the student and teacher's features. This suggests redundancy in explicitly constructing and updating a large memory bank of previous representations~\cite{Tian2019ContrastiveDistillation}. By considering a weight decay $\eta$ and a learning rate $\alpha_p$, the update equation can be given as follows:
\vspace{-0.7em}

\begin{align}
    \mathbf{W}_p &\rightarrow \mathbf{W}_p + \alpha_p\dot{\mathbf{W}_p} - \eta\mathbf{W}_p \\
    &= (1 - \eta)\mathbf{W}_p + \alpha_p\dot{\mathbf{W}_p}
\end{align}

By setting $\eta = \alpha_p$ we can see that the projection layer will reduce to a moving average of relational features, which is very similar to the momentum encoder used by CRD~\cite{Tian2019ContrastiveDistillation}. Other works suggest to extract relational information on-the-fly by constructing correlation or Gram matrices~\cite{Miles2022InformationDistillation, Peng2019CorrelationDistillation}. We show that this is also not necessary and more complex information can be captured through a simple linear projector. We also demonstrate that, in general, the use of a projector will scale much more favourably for larger batch sizes and feature dimensions. We also note that the handcrafted design of kernel functions~\cite{Joshi2021OnNetworks, He2022FeatureDistillation} may not generalise to real large scale datasets without significant hyperparameter tuning.

From the results in table \ref{table:rebuttal_normalisation}, we observe that when fixing all other settings, the choice of normalisation can significantly affect the student's performance. To explore this in more detail, we consider the training trajectories of $\mathbf{W}_p$ under  different normalisation schemes. We find that the choice of normalisation not only controls the training dynamics, but also the fixed point solution (see equation \ref{eqn:projector_update}). We argue that the efficacy of distillation is dependent on how much relational information can be encoded in the learned weights and how much information is lost through the projection. To jointly evaluate these two properties we show the evolution of singular values of the projector weights during training. The results can be seen in figure \ref{fig:singular_values_evolution} and show that the better performing normalisation methods (table \ref{table:rebuttal_normalisation}) are shrinking far fewer singular values towards zero. This shrinkage can be described as collapsing the input along some dimension, which will induce some information loss and it is this information loss that degenerates the efficacy of the distillation process.

\begin{table}[H]
    \centering
    \small
    \input{tables/rebuttal/normalisation}
    \caption{Normalisation ablation for distillation across a range of architecture pairs on ImageNet-1K 20\% subset.}
    \label{table:rebuttal_normalisation}
\end{table}

\begin{figure}[H]
\centering
\includegraphics[width=.86\linewidth]{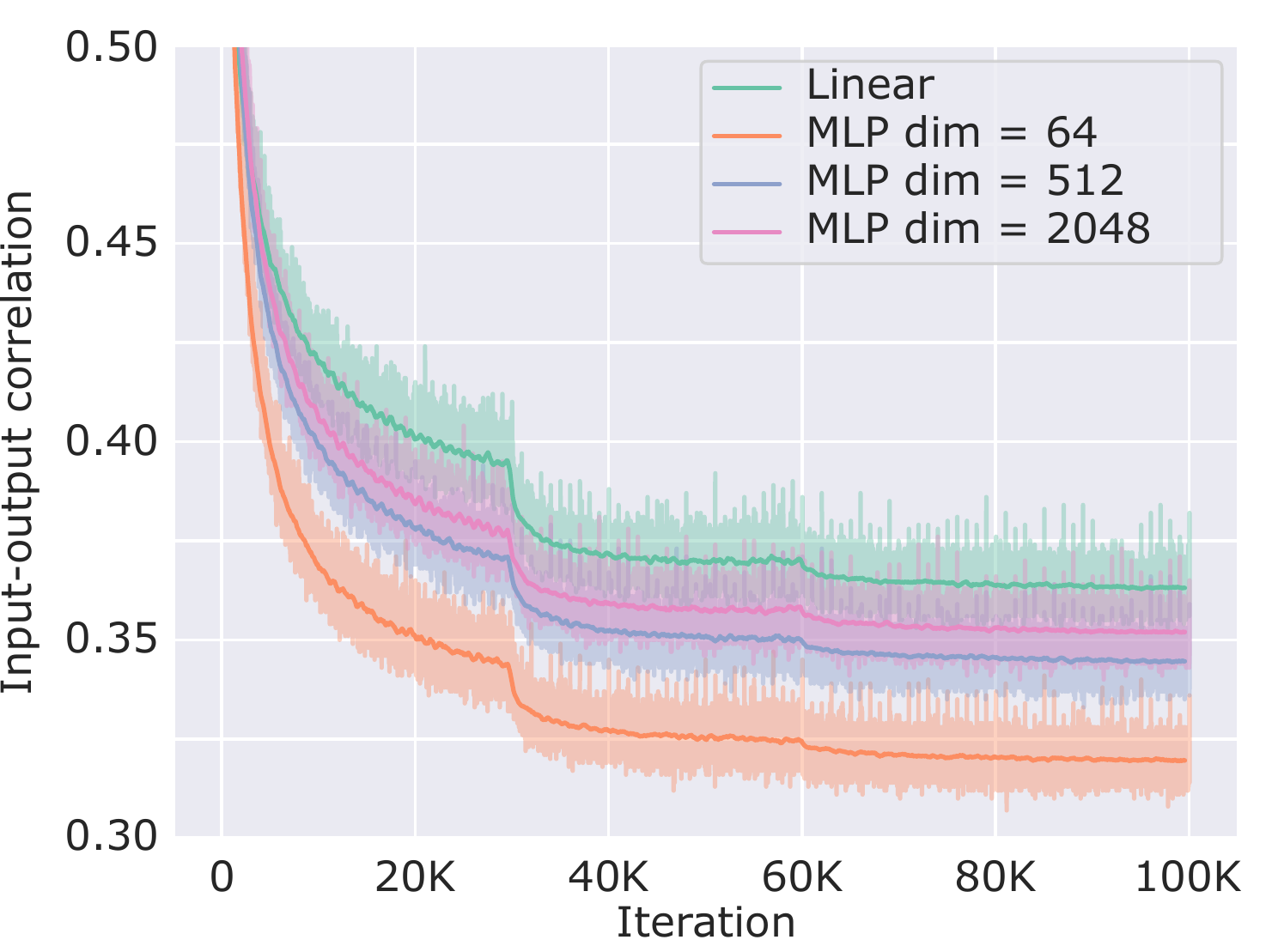}
\vspace{-0.7em}
\caption{Correlation between input-output features using different projector architectures. All projectors considered will gradually decorrelate the input-output features.}
\label{fig:input_output_decorrelation}
\end{figure}

\textbf{Larger projector networks learn to decorrelate the input-output features.}
One natural extension of the previous observations is to use a larger projector network to encode more information relevant for the distillation loss. Unfortunately, we observe that a trivial expansion of the projection architecture does not necessarily improve the students performance. To explain this observation we evaluate a measure of decorrelation between the input and output features of these projector networks. The results can be seen in figure \ref{fig:input_output_decorrelation} and we can see that the larger projectors learn to decorrelate more and more features from the input. This decorrelation can lead to the projector learning features that are not shared with the student backbone, which will subsequently diminish the effectiveness of distillation. These observations suggest that there is an inherent trade-off between the projector capacity and the efficacy of distillation. We note that the handcrafted design of the projector architecture is a motivated direction for further research~\cite{Chen2022ImprovedEnsemble, Navaneet2021SimReg:Distillation}.
However, in favour of simplicity, we choose to use a linear projector for all of our large scale evaluations. 

\textbf{The soft maximum function can address distilling across a large capacity gap.}
 When the capacity gap between the student and the teacher is large, representation distillation can become challenging. More specifically, the student network may have insufficient capacity to perfectly align these two spaces and in attempting to do so may degrade its downstream performance. To addresses this issue we explore the use of a soft maximum function which will soften the contribution of relatively close matches in a batch. In this way the loss can be adjusted to compensate for poorly aligned features which may arise when the capacity gap is large. The family of functions which share these properties can be more broadly defined through a property of their gradients. In favour of simplicity, we use the simple $LogSum$ function throughout our experiments.

 \begin{align} 
     D(\mathbf{Z}_s, \mathbf{Z}_t \;; \mathbf{W}_p) = log \sum_i \lvert\mathbf{Z}_s\mathbf{W}_p - \mathbf{Z}_t\rvert_i^{\alpha}
     \label{eqn:dlogsum}
\end{align}

where $\alpha$ is a smoothing factor. We also note that other functions, such as the $LogSumExp$, with a temperature parameter $\tau$, have been used in SimCLR and CRD to a similar effect. Table~\ref{table:rebuttal_normalisation} shows the importance of feature normalisation across a variety of student-teacher architecture pairs. Batch normalisation provides the most consistent improvement that even extends to the Transformer $\rightarrow$ CNN setting. In table~\ref{table:logsum} we highlight the importance of the $LogSum$ function, which is most effective in the large capacity gap settings, as evident from the 1\% improvement for R50 $\rightarrow$ R18. Table~\ref{table:alpha} provides an ablation of the importance of the $\alpha$ parameter, whereby we observe that the performance is relatively robust to a wide range of values, but consistently optimal in the range 4-5.

\begin{table}[t!]
    \centering
    \small
\input{tables/rebuttal/logsum}
    \caption{LogSum ablation across various architecture pairs. Left: 20\% subset. Right: Full ImageNet. The soft maximum function provides consistent improvement across both the CNN$\rightarrow$CNN and ViT$\rightarrow$CNN distillation settings.} 
    \label{table:logsum}
\end{table}

\begin{table}[H]
    \centering
    \small
\input{tables/rebuttal/alpha_new2}
    \caption{Ablating the importance of $\alpha$. Distillation is generally robust for various values of $\alpha$, but consistently optimal in range 4-5 across various architecture pairs.}
    \label{table:alpha}
\end{table}

% main experiments
\section{Benchmark Evaluation}
\label{sec:benchmark_evaluation}

\textbf{Implementation details.}
We follow the same training schedule as CRD~\cite{Tian2019ContrastiveDistillation} for both the CIFAR100 and ImageNet experiments. For the object detection, we use the same training schedule as ReviewKD~\cite{Chen2021DistillingReview}, while for the data efficient training we use the same as Co-Advice~\cite{Ren2022Co-advise:Distillation}. All experiments were performed on a single NVIDIA RTX A5000. When using batch normalisation for the representations, we removed the affine parameters and set $\epsilon = 0.0001$. For all experiments we jointly train the student using a task loss $\mathcal{L}_{task}$ and the feature distillation loss given in equation \ref{eqn:dlogsum}.

\begin{align}
    \mathcal{L} = \mathcal{L}_{task} + D(\mathbf{Z}_s, \mathbf{Z}_t \;; \mathbf{W}_p)
\end{align}

% An additional logit distillation loss is added to some experiments to improve the student performance.

\subsection{Data Efficient Training for Transformers}
\label{sec:deit}
Transformers have emerged as a viable replacement for convolution-based neural networks (CNN) in visual learning tasks. Despite the promise of these models, their performance will suffer when there is insufficient training data available, such as in the case of ImageNet. DeiT~\cite{Touvron2021TrainingAttention} was the first to address this problem through the use of knowledge distillation. Although the authors show improved alignment with the teacher, we believe this fails to capture why less data is needed.

\begin{table}[H]
    \centering
    \small
    \input{tables/deit_new2.tex}
    \caption{Data-efficient training of transformers and CNNs on the ImageNet-1K dataset. Unless specified, all student models are trained for 300 epochs.}
    \vspace{-0.5em}
    \label{table:data_efficient_training_transformers}
\end{table}

\begin{table*}[ht!]
    \centering
    \small
    \input{tables/cifar100.tex}
    \caption{KD between Similar and Different Architectures. Top-1 accuracy (\%) on CIFAR100. Bold is used to denote the best results. All reported models are trained using pairs of augmented images. Those reported in the top box use RandAugment~\cite{Cubuk2020Randaugment:Space} strategy, while those in the bottom box use pre-defined rotations, as used in SSKD. $^\dagger$ denotes reproduced results in a new augmentation setting using the authors provided code.}
    \label{table:cifar100}
    \vspace{-0.7em}
\end{table*}

We posit that the distillation process encourages the student to learn layers which are "more" translational equivariant in attempt to match the teacher's underlying function. Although this is the principle that motivates using an ensemble of teacher models with different inductive biases~\cite{Ren2022Co-advise:Distillation}, there is still no thorough demonstration on if the inductive biases are actually being transferred. In this section we attempt to address this gap by introducing a measure of equivariance. We show that applying our distillation principles to this task can achieve significant improvements over state-of-the-art as a result of transferring more of the translational equivariance.

The results of these experiments are shown in table \ref{table:data_efficient_training_transformers}. We use the exact same training methodology as co-advice~\cite{Ren2022Co-advise:Distillation} and choose to use batch normalisation, a linear projection layer, and $\alpha = 4$ as the parameters for distillation. We observe a significant improvement over both DeiT and CivT when the capacity gap is large. However, as the capacity gap diminishes, and the student approaches the same performance as the teacher, this improvement is much less significant. Multiple factors, such as the soft maximum function and the batch normalisation, will be contributing to this observed result. However, the explanation is more concisely described by the fact that our distillation loss transfers more translational equivariance to the student.

\subsection{Classification on CIFAR100 and ImageNet}
\label{sec:cifar_and_imagenet}
Experiments on the CIFAR-100 classification task~\cite{Krizhevsky2009LearningImages} consist of 60K 32×32 RGB images across
100 classes with a 5:1 training/testing split. Table \ref{table:cifar100} shows the results for several student-teacher pairings. To enable a fair evaluation, we have only included the methods that use the same teacher weights provided by SSKD\cite{Xu2020KnowledgeSelf-supervision}. In these experiments we use an MLP projector with a hidden size of 1024 and no additional KL divergence loss. We confirm that not only is the choice of augmentation critical for good performance~\cite{Beyer2022KnowledgeConsistent} on this dataset, but applying our principles can attain state-of-the-art across most architecture pairs. The most significant improvements pertain to the cross-architecture experiments or where the capacity gap is large. We provide two sets of experiments with and without introducing a wider set of augmentations. In both settings we maintain the same optimiser, learning rate scheduler, and training duration.

The ImageNet~\cite{Russakovsky2014ImageNetChallenge} classification uses 1.3 million images that are classified into 1000 distinct classes. The input size are set to 224 x 224, and we employed a typical augmentation procedure that includes cropping and horizontal flipping. We used the torchdistill library with the standard configuration, which involves 100 training epochs using SGD and an initial learning rate of 0.1, which is decreased by a factor of 10 at epochs 30, 60, and 90. The results can be seen in table \ref{table:imagenet} and although the choice of architectures is not in favour of our method since the capacity gap is small, we are still able to attain competitive performance. Other methods, such as ICKD~\cite{Liu2021ExploringDistillation} or SimKD~\cite{Chen2022KnowledgeClassifier} either modify the original training settings or architectures, and so have been omitted from this evaluation.

\paragraph{Cross architecture distillation can implicitly transfer inductive biases.} CNNs use convolutions, which are spatially local operations, whereas transformers use self-attention, which are global operations. We expect that a benefit of this cross-architecture distillation setting is that the students learn to be "more" spatially equivariant in an attempt to match the teachers underlying function. It is this strong inductive bias that can reduce the amount of training data needed. A layer is translation equivariant if the following property holds:

\begin{align}
    \phi(T\mathbf{x}) = T\phi(\mathbf{x})
    \label{eqn:equivariance}
\end{align}

In other words, if we take a translated input $T\mathbf{x}$ and pass it through a layer $\phi$, the result should be equivalent to first applying $\phi$ to $\mathbf{x}$ and then performing the translation. A natural measure of equivariance can then be the difference between the left and right-hand side of this equation \ref{eqn:equivariance}.

\begin{align}
    \mu_{T}(\phi) = \lpnorm{\phi(T\mathbf{x}) - T\phi(\mathbf{x})}_2^2
\end{align}

\begin{table*}[htp]
    \centering
    \small
    \input{tables/imagenet.tex}
    \vspace{-.8em}
    \caption{Top-1 and Top-5 error rates (\%) on ImageNet. ResNet18 as student, ResNet34 as teacher.}
    \label{table:imagenet}
\end{table*}

We evaluate this measure on a block of self-attention layers by first removing the distillation and class tokens and then rolling the patch tokens to recover the spatial dimensions. This operation can then be performed on the input and output tensors before applying a translation. Table \ref{table:equivariance_measure} shows this measure of equivariance after training with and without distillation. In general, we observe that the distilled models do in fact learn to preserve spatial locality between feature maps, which aligns with the function matching perspective for distillation. 

We find that our simple distillation recipe can transfer a lot more of this equivariance property to the student. Although Co-advise does learn this spatial locality to some extent, it is much less significant than using our feature based distillation, despite both attaining a similar level of performance. Intermediate feature map losses may be able to transfer even more of this translational equivariance property, however, its usage may degrade the benefit of using a transformer in the first place. For example, although we observe that most self-attention blocks (trained using distillation) do preserve a lot of this spatial locality, there is still some global context between patch tokens that is still being preserved.

\begin{table}[H]
    \centering
    \small
    \input{tables/translation_equivariance}
    \caption{Measure of translational equivariance for a DeiT-S transformer model trained with and without distillation. These results confirm that distillation can transfer explicit inductive biases from the teacher.}
    % Reported measure was computed after every epoch and averaged over 256 images.
    \label{table:equivariance_measure}
\end{table}

\subsection{Object Detection on COCO}
\label{sec:object_detection}

We extend the application of our method to object detection, whereby we employ a similar approach as used in the classification task by distilling the backbone output features of both the student and teacher networks. Due to the smaller batch sizes used in these experiments, we choose to instead normalise over the height and width dimensions. For evaluating the efficacy of our method, we conduct experiments on the widely-used COCO2017 dataset~\cite{Lin2014MicrosoftContext} under the same settings provided in ReviewKD. We then further demonstrate the applicability of our distillation principles on the more recent and efficient YOLOv5 model~\cite{Zhu2021TPH-YOLOv5:Scenarios}. In both cases we show improved student performance on the downstream task, whereby competitive performance is achieved with ReviewKD despite being significantly simpler and cheaper to integrate into a given distillation pipeline. Our method even outperforms FPGI~\cite{Wang2019DistillingImitationb}, which is directly designed for detection.

% (top) Single-model single-scale mAP on COCO val2017. All models are trained for 300 epochs.
\begin{table}
    \centering
    \small
    \input{tables/yolo_no_flops_params}
    \caption{Object detection on COCO. (top) We report the standard COCO metric of mAP averaged over IOU thresholds in [0.5 : 0.05 : 0.95] along with the standard PASCAL VOC’s metric~\cite{pascalvoc}, which is the average mAP@0.5. (bottom) For the R-CNN results, we report the mAP and AP50 metrics to enable a consistent comparison with ReviewKD.}
    \label{table:yolo}
\end{table} 

% conclusion
\section{Conclusion}
\label{sec:conclusion}
In this paper, we revisited the core underlying principles of knowledge distillation and have performed an extensive ablation on the most effective and scaleable components. In doing so, we have provided a new theoretical perspective for understanding these results through analyzing the projector training dynamics. By extending these principles to a wide range of tasks, we achieve competitive or improved performance to state-of-the-art across image classification, object detection, and data efficient training of transformers. Our proposed distillation recipe can significantly reduce the complexity and memory consumption of existing pipelines by avoiding the need to construct expensive relational object, many trainable layers, or enforcing very long training schedules. We further show improved performance for the large capacity gap settings and evidence for the distillation of explicit inductive biases from the teacher. Looking ahead to future research in this area, we expect to see the joint development of more sophisticated normalisation schemes and projection networks, which will encode more complex and informative features for the distillation process. 

\paragraph{Code Reproducibility.} To facilitate the reproducibility of results, we release all the training code and pre-trained weights. The ImageNet experiments are performed using the popular \textit{torchdistill}~\cite{Matsubara2020TorchdistillDistillation} framework, while the CIFAR100 and data-efficient training code are based on those provided by CRD~\cite{Tian2019ContrastiveDistillation} and co-advice~\cite{Ren2022Co-advise:Distillation} respectively.

% \section{References}
% \label{sec:references}
% \printbibliography
% \bibliography{references, more_references}
\bibliography{main}

% {\small
% \bibliographystyle{ieee_fullname}
% \bibliography{references}
% }

% \input{supplementary}

\end{document}

%% file: tables/rebuttal/normalisation.tex
\begin{tabular}{lccc}
    \toprule
    \belowrulesepcolor{mygrey}
    \rowcolor{mygrey} Teacher & RegNet & ViT & ConvNeXt \\
    \rowcolor{mygrey} Student & MBv2 & MBv3 & EffNet-b0 \\
    \aboverulesepcolor{mygrey}
    \midrule
    No distillation & 50.89 & 54.13 & 64.48 \\
    L2 Norm & 52.91 & 54.65 & 64.23 \\
    Group Norm & 55.63 & 59.08 & 65.52 \\
    Batch Norm & \textbf{56.09} & \textbf{59.28} & \textbf{67.95} \\
    \midrule  
\end{tabular}

%% file: tables/rebuttal/logsum.tex
\begin{tabular}{lcc|c}
    \toprule
    \belowrulesepcolor{mygrey}
    \rowcolor{mygrey} Teacher & ViT & ConvNeXt & ResNet50 \\
    \rowcolor{mygrey} Student & MBv3 & EffNet-b0 & ResNet18 \\
    \aboverulesepcolor{mygrey}
    \midrule
    wo/ LogSum & 59.28 & 67.95 & 70.03 \\
    w/ LogSum & \textbf{59.80} & \textbf{68.51} & \textbf{71.29} \\
    \midrule
\end{tabular}

%% file: tables/rebuttal/alpha_new2.tex
\begin{tabular}{cccccc}
    \toprule
    \belowrulesepcolor{mygrey}
    \rowcolor{mygrey} Teacher & ResNet50 & ConvNeXt \\
    \rowcolor{mygrey} Student & ResNet18 & EffNet-b0 \\
    \aboverulesepcolor{mygrey}
    \midrule
    1.0 & 61.74 & 65.52 \\
    2.0 & 62.23 & 66.61 \\
    3.0 & 63.10 & 67.72 \\
    4.0 & 63.32 & \textbf{68.51} \\
    5.0 & \textbf{63.40} & 67.69 \\
    \midrule
\end{tabular}

%% file: tables/deit_new2.tex
\begin{tabular}{lccc}
    
    \toprule
    \belowrulesepcolor{mygrey} 
    \rowcolor{mygrey} Network & acc@1 & Teacher & \#params \\
    \aboverulesepcolor{mygrey}
    \midrule
    RegNetY-160 & 82.6 & none & 84M \\
    % 236,335,208
    BiT-M R152x2 & 84.5 & none & 236M \\
    % ~\cite{Radosavovic2020DesigningSpaces}
    % CaiT-S24~\cite{touvron2021going} & 83.4 & none & 47M \\
    % DeiT3-B~\cite{touvron2022deit} & \textbf{83.8} & none & 87M \\
    % CeiT-S~\cite{yuan2021incorporating} \textit{\footnotesize ICCV21} & 83.3 & none & 24M \\
    % ResNet101 & - & none & -M \\
    % \midrule
    % ViT-B~\cite{dosovitskiy2021an} \textit{\footnotesize ICLR21} & 77.9 & none & 86M \\
    % ViT-L~\cite{dosovitskiy2021an} \textit{\footnotesize ICLR21} & 76.5 & none & 307M \\
    \midrule
    DeiT-Ti & 72.2 & none & 5M\\
    CivT-Ti & 74.9 & ensemble & 6M \\
    % Closer Look~\cite{Miles2023ADistillation} & 77.2 & regnet-y-160 & 6M \\
    % TRG~\cite{zhang2023knowledge} \textit{\footnotesize TBDATA23} & 75.5 & ceit-s & 6M \\
    % \emoji{angry} ViTKD~\cite{yang2022vitkd} & 77.8 & regnet-y-160 & 6M \\
    % Manifold~\cite{hao2022learning} & 76.5 & cait-s24 & 6M \\
    DeiT-Ti\alambicdeit & 74.5 & regnety-160 & 6M \\
    \myalign{l}{\;\;\;\footnotesize \rotatebox[origin=c]{180}{$\Lsh$} 1000 epochs } & 76.6 & regnety-160 & 6M \\
    DearKD & 74.8 & regnety-160 & 6M \\
    \myalign{l}{\;\;\;\footnotesize \rotatebox[origin=c]{180}{$\Lsh$} 1000 epochs } & 77.0 & regnety-160 & 6M \\
    USKD & 75.0 & regnety-160 & 6M \\
    Our Method & \textbf{77.2} & regnety-160 & 6M \\
    \midrule
    
    ResNet-50 & 76.5 & none & 25M \\
    FunMatch & 80.3 & bit-m r152x2 & 25M \\
    $\rotatebox[origin=c]{180}{$\Lsh$}$ 9600 epochs & \textbf{82.8} & bit-m r152x2 & 25M \\
    
    \midrule
    DeiT-S & 79.8 & none & 22M\\
    CivT-S & 82.0 & ensemble & 22M \\
    DeiT-S\alambicdeit & 81.2 & regnety-160 & 22M \\
    \myalign{l}{\;\;\;\footnotesize \rotatebox[origin=c]{180}{$\Lsh$} 1000 epochs } & 82.6 & regnety-160 & 22M \\
    DearKD & 81.5 & regnety-160 & 22M \\
    \myalign{l}{\;\;\;\footnotesize \rotatebox[origin=c]{180}{$\Lsh$} 1000 epochs } & 82.8 & regnety-160 & 22M \\  
    USKD & 80.8 & regnety-160 & 22M \\
    Our Method & \textbf{82.1} & regnety-160 & 22M \\
    \bottomrule
\end{tabular}

%% file: tables/cifar100.tex
\begin{tabular}{lcccc|ccccccc}
    \toprule

    \belowrulesepcolor{mygrey} 
    \rowcolor{mygrey} Teacher & WRN40-2 & WRN40-2 & R56 & R32×4 & VGG13 & R50 & R50 & R32×4 & R32×4 & WRN40-2\\
    \rowcolor{mygrey} Student & WRN16-2 & WRN40-1 & R20 & R8×4 & MBv2 & MBv2 & VGG8 & ShuffleV1 & ShuffleV2 & ShuffleV1 \\
    \aboverulesepcolor{mygrey}
    \midrule
    Teacher & 76.46 & 76.46 & 73.44 & 79.63 & 75.38 & 79.10 & 79.10 & 79.63 & 79.63 & 76.46 \\
    Student & 73.64 & 72.24 & 69.63 & 72.51 & 65.79 & 65.79 & 70.68 & 70.77 & 73.12 & 70.77 \\
    \midrule
    KD & 74.92 & 73.54 & 70.66 & 73.33 & 67.37 & 67.35 & 73.81 & 74.07 & 74.45 & 74.83 \\
    FitNet & 75.75 & 74.12 & 71.60 & 74.31 & 68.58 & 68.54 & 73.84 & 74.82 & 75.11 & 75.55\\
    AT & 75.28 & 74.45 & 71.78 & 74.26 & 69.34 & 69.28 & 73.45 & 74.76 & 75.30 & 75.61\\
    CRD & 76.04 & 75.52 & 71.68 & 75.90 & 68.49 & 70.32 & 74.42 & 75.46 & 75.72 & 75.96\\
    SSKD & 76.04 & \textbf{76.13} & 71.49 & 76.20 & 71.53 & 72.57 & 75.76 & \textbf{78.44} & 78.61 & 77.40 \\

    Our Method & \textbf{76.14} & 75.42 & \textbf{71.75} & \textbf{76.44} & \textbf{71.47} & \textbf{72.81} & \textbf{76.20} & 77.32 & \textbf{79.06} & \tabnode{\textbf{79.22}} \\
    \midrule

    % & & & & & \\[-.5ex]
    KD$^\dagger$ & 75.94 & 75.32 & 71.10 & 75.84 & 70.79 & 71.29 & 75.75 & 77.80 & 78.43 & 78.00 \\
    CRD$^\dagger$ & 77.27 & \textbf{76.15} & 72.21 & 77.69 & 71.65 & 72.03 & 75.73 & 78.57 & 79.01 & 78.54 \\
    DKD$^\dagger$ & 74.96 & 75.89 & 70.95 & 77.52 & 72.01 & 73.30 & 76.88 & \textbf{79.71} & \textbf{80.08} & 77.86 \\

    Our Method & \textbf{77.61} & 76.04 & \textbf{72.25} & \textbf{78.37} & \textbf{72.82} & \textbf{73.51} & \textbf{77.08} & 78.99 & 79.86 & \tabnode{\textbf{78.79}} \\
    \bottomrule
    
\end{tabular}

%% file: tables/imagenet.tex
\begin{tabular}{c|cc|ccccc|c}
    \toprule
    
    % & Teacher & Student & AT~\cite{Zagoruyko2019PayingTransfer} & KD~\cite{Hinton2015DistillingNetwork} & SP~\cite{Tung2019Similarity-preservingDistillation} & CC~\cite{Peng2019CorrelationDistillation} & CRD~\cite{Hinton2015DistillingNetwork} & ReviewKD~\cite{Chen2021DistillingReview} & 
    % Ours \\
    \belowrulesepcolor{mygrey}
    \rowcolor{mygrey} & Teacher & Student & AT & KD & CC & CRD & ReviewKD & Ours \\
    \aboverulesepcolor{mygrey}
    % & & & \cite{Zagoruyko2019PayingTransfer} & \cite{Hinton2015DistillingNetwork} & \cite{Peng2019CorrelationDistillation} & \cite{Tian2019ContrastiveDistillation} & \cite{Chen2021DistillingReview} & \\
    \midrule
    acc@1 & 26.69 & 30.25 & 29.30 & 29.34 & 30.04 & 28.62 & 28.39 & \textbf{28.37} \\
    acc@5 & 8.58 & 10.93 & 10.00 & 10.12 & 10.83 & 9.51 & 9.42 & \textbf{9.41} \\

    % \multirow{2}{*}{(b)} & acc@1 & - & - & - & - & - & - & - & - & - \\
    % & acc@5 & - & - & - & - & - & - & - & - & - \\
    
    \midrule
\end{tabular}

%% file: tables/translation_equivariance.tex
\begin{tabular}{lc}
    
    \toprule
    \belowrulesepcolor{mygrey} 
    \rowcolor{mygrey} Network & $\mu_T(\phi)$	\\
    \aboverulesepcolor{mygrey}
    \midrule
    DeiT-S & $1.52 \pm 0.15$ \\
    CivT-S & $0.13 \pm 0.05$ \\
    Our Method & $\bf{0.04 \pm 0.02}$ \\
    \bottomrule
\end{tabular}

%% file: tables/yolo_no_flops_params.tex
\begin{tabular}{lcc}
    
    \toprule
    \belowrulesepcolor{mygrey} 
    \rowcolor{mygrey} Model & mAP (50-95) & mAP 50 \\
    \aboverulesepcolor{mygrey}
    \midrule
    YOLOv5m (teacher) & 64.1 & 45.4 \\
    % \midrule
    YOLOv5s & 56.8 & 37.4 \\

    $\rotatebox[origin=c]{180}{$\Lsh$}$ Our Method & \textbf{57.3} & \textbf{37.5} \\
    
    \midrule
    % \belowrulesepcolor{mygrey} 
    & mAP & AP50 \\
    % \aboverulesepcolor{mygrey}
    \midrule
    
    Faster R-CNN w/ R50 (teacher) & 40.22 & 61.02 \\
    % \midrule
    Faster R-CNN w/ MV2 & 29.47 & 48.87 \\
    $\rotatebox[origin=c]{180}{$\Lsh$}$ KD & 30.13 & 50.28 \\
    $\rotatebox[origin=c]{180}{$\Lsh$}$ FitNet & 30.20 & 49.80 \\
    $\rotatebox[origin=c]{180}{$\Lsh$}$ FPGI & 31.16 & 50.68 \\
    $\rotatebox[origin=c]{180}{$\Lsh$}$ ReviewKD & \textbf{33.71} & \textbf{53.15} \\
    % epoch 299/300
    $\rotatebox[origin=c]{180}{$\Lsh$}$ Our Method & \underline{32.92} & \underline{52.96} \\

    \bottomrule
    
\end{tabular}

%% file: main.bbl
\begin{thebibliography}{61}
\providecommand{\natexlab}[1]{#1}

\bibitem[{Allen-Zhu and Li(2023)}]{allen-zhu2023towards}
Allen-Zhu, Z.; and Li, Y. 2023.
\newblock Towards Understanding Ensemble, Knowledge Distillation and Self-Distillation in Deep Learning.

\bibitem[{Bardes, Ponce, and LeCun(2022{\natexlab{a}})}]{Bardes2022VICReg:Learning}
Bardes, A.; Ponce, J.; and LeCun, Y. 2022{\natexlab{a}}.
\newblock {VICReg: Variance-Invariance-Covariance Regularization for Self-Supervised Learning}.
\newblock \emph{ICLR}.

\bibitem[{Bardes, Ponce, and LeCun(2022{\natexlab{b}})}]{Bardes2022VICRegL:Features}
Bardes, A.; Ponce, J.; and LeCun, Y. 2022{\natexlab{b}}.
\newblock {VICRegL: Self-Supervised Learning of Local Visual Features}.
\newblock \emph{NeurIPS}.

\bibitem[{Beyer et~al.(2022)Beyer, Zhai, Royer, Markeeva, Anil, and Kolesnikov}]{Beyer2022KnowledgeConsistent}
Beyer, L.; Zhai, X.; Royer, A.; Markeeva, L.; Anil, R.; and Kolesnikov, A. 2022.
\newblock {Knowledge distillation: A good teacher is patient and consistent}.
\newblock \emph{CVPR}.

\bibitem[{Carlucci et~al.(2019)Carlucci, D’Innocente, Bucci, Caputo, and Tommasi}]{Carlucci2019DomainPuzzles}
Carlucci, F.~M.; D’Innocente, A.; Bucci, S.; Caputo, B.; and Tommasi, T. 2019.
\newblock {Domain Generalization by Solving Jigsaw Puzzles}.
\newblock \emph{CVPR}.

\bibitem[{Chen et~al.(2022{\natexlab{a}})Chen, Mei, Zhang, Wang, Feng, and Chen}]{Chen2022KnowledgeClassifier}
Chen, D.; Mei, J.-P.; Zhang, H.; Wang, C.; Feng, Y.; and Chen, C. 2022{\natexlab{a}}.
\newblock {Knowledge Distillation with the Reused Teacher Classifier}.
\newblock \emph{CVPR}.

\bibitem[{Chen et~al.(2020{\natexlab{a}})Chen, Wang, Gan, Liu, Henao, and Carin}]{Chen2020WassersteinDistillation}
Chen, L.; Wang, D.; Gan, Z.; Liu, J.; Henao, R.; and Carin, L. 2020{\natexlab{a}}.
\newblock {Wasserstein Contrastive Representation Distillation}.
\newblock \emph{CVPR}.

\bibitem[{Chen et~al.(2021{\natexlab{a}})Chen, Liu, Zhao, and Jia}]{Chen2021DistillingReview}
Chen, P.; Liu, S.; Zhao, H.; and Jia, J. 2021{\natexlab{a}}.
\newblock {Distilling Knowledge via Knowledge Review}.
\newblock \emph{CVPR}.

\bibitem[{Chen et~al.(2020{\natexlab{b}})Chen, Kornblith, Norouzi, and Hinton}]{Chen2020ARepresentations}
Chen, T.; Kornblith, S.; Norouzi, M.; and Hinton, G. 2020{\natexlab{b}}.
\newblock {A simple framework for contrastive learning of visual representations}.
\newblock \emph{ICML}.

\bibitem[{Chen et~al.(2020{\natexlab{c}})Chen, Fan, Girshick, and He}]{Chen2020ImprovedLearning}
Chen, X.; Fan, H.; Girshick, R.; and He, K. 2020{\natexlab{c}}.
\newblock {Improved Baselines with Momentum Contrastive Learning}.
\newblock \emph{arXiv preprint}.

\bibitem[{Chen, Xie, and He(2021)}]{Chen2021AnTransformers}
Chen, X.; Xie, S.; and He, K. 2021.
\newblock {An Empirical Study of Training Self-Supervised Vision Transformers}.
\newblock \emph{ICCV}.

\bibitem[{Chen et~al.(2021{\natexlab{b}})Chen, Bian, Xiao, Rong, Xu, and Huang}]{chen2021selfdistilling}
Chen, Y.; Bian, Y.; Xiao, X.; Rong, Y.; Xu, T.; and Huang, J. 2021{\natexlab{b}}.
\newblock On Self-Distilling Graph Neural Network.

\bibitem[{Chen et~al.(2022{\natexlab{b}})Chen, Wang, Liu, Xu, de~Hoog, and Huang}]{Chen2022ImprovedEnsemble}
Chen, Y.; Wang, S.; Liu, J.; Xu, X.; de~Hoog, F.; and Huang, Z. 2022{\natexlab{b}}.
\newblock {Improved Feature Distillation via Projector Ensemble}.
\newblock \emph{NeurIPS}.

\bibitem[{Cubuk et~al.(2020)Cubuk, Zoph, Shlens, and Le}]{Cubuk2020Randaugment:Space}
Cubuk, E.~D.; Zoph, B.; Shlens, J.; and Le, Q.~V. 2020.
\newblock {Randaugment: Practical automated data augmentation with a reduced search space}.
\newblock \emph{CVPR Workshop}.

\bibitem[{Doersch, Gupta, and Efros(2015)}]{Doersch2015UnsupervisedPrediction}
Doersch, C.; Gupta, A.; and Efros, A.~A. 2015.
\newblock {Unsupervised Visual Representation Learning by Context Prediction}.
\newblock \emph{ICCV}.

\bibitem[{Ermolov et~al.(2020)Ermolov, Siarohin, Sangineto, and Sebe}]{Ermolov2020WhiteningLearning}
Ermolov, A.; Siarohin, A.; Sangineto, E.; and Sebe, N. 2020.
\newblock {Whitening for Self-Supervised Representation Learning}.
\newblock \emph{ICML}.

\bibitem[{Everingham et~al.(2010)Everingham, Gool, Williams, Winn, and Zisserman.}]{pascalvoc}
Everingham, M.; Gool, L.~V.; Williams, C. K.~I.; Winn, J.; and Zisserman., A. 2010.
\newblock The pascal visual object classes (voc) challenge.

\bibitem[{Gidaris, Singh, and Komodakis(2018)}]{Gidaris2018UnsupervisedRotations}
Gidaris, S.; Singh, P.; and Komodakis, N. 2018.
\newblock {Unsupervised Representation Learning by Predicting Image Rotations}.
\newblock \emph{ICLR}.

\bibitem[{Guo et~al.(2020)Guo, Chen, Hu, Zhu, He, and Cai}]{Guo2020ReducingDistillation}
Guo, J.; Chen, M.; Hu, Y.; Zhu, C.; He, X.; and Cai, D. 2020.
\newblock {Reducing the Teacher-Student Gap via Spherical Knowledge Distillation}.
\newblock \emph{arXiv preprint}.

\bibitem[{He and Ozay(2022)}]{He2022FeatureDistillation}
He, B.; and Ozay, M. 2022.
\newblock {Feature Kernel Distillation}.
\newblock \emph{ICLR}.

\bibitem[{He et~al.(2022)He, Chen, Xie, Li, Doll{\'{a}}r, and Girshick}]{He2022MaskedLearners}
He, K.; Chen, X.; Xie, S.; Li, Y.; Doll{\'{a}}r, P.; and Girshick, R. 2022.
\newblock {Masked Autoencoders Are Scalable Vision Learners}.
\newblock \emph{CVPR}.

\bibitem[{He et~al.(2020)He, Fan, Wu, Xie, and Girshick}]{He2020MomentumLearning}
He, K.; Fan, H.; Wu, Y.; Xie, S.; and Girshick, R. 2020.
\newblock {Momentum Contrast for Unsupervised Visual Representation Learning}.
\newblock \emph{CVPR}.

\bibitem[{Hinton, Vinyals, and Dean(2015)}]{Hinton2015DistillingNetwork}
Hinton, G.; Vinyals, O.; and Dean, J. 2015.
\newblock {Distilling the Knowledge in a Neural Network}.
\newblock \emph{NeurIPS}.

\bibitem[{Huang and Wang(2017)}]{Huang2017LikeTransfer}
Huang, Z.; and Wang, N. 2017.
\newblock {Like What You Like: Knowledge Distill via Neuron Selectivity Transfer}.
\newblock \emph{arXiv preprint}.

\bibitem[{Joshi et~al.(2021)Joshi, Liu, Xun, Lin, and Foo}]{Joshi2021OnNetworks}
Joshi, C.~K.; Liu, F.; Xun, X.; Lin, J.; and Foo, C.-S. 2021.
\newblock {On Representation Knowledge Distillation for Graph Neural Networks}.
\newblock \emph{arXiv preprint}.

\bibitem[{Krizhevsky(2009)}]{Krizhevsky2009LearningImages}
Krizhevsky, A. 2009.
\newblock {Learning Multiple Layers of Features from Tiny Images}.

\bibitem[{Krizhevsky, Sutskever, and Hinton(2012)}]{Krizhevsky2012ImageNetNetworks}
Krizhevsky, A.; Sutskever, I.; and Hinton, G.~E. 2012.
\newblock {ImageNet Classification with Deep Convolutional Neural Networks}.
\newblock \emph{NeurIPS}.

\bibitem[{Lin et~al.(2014)Lin, Maire, Belongie, Hays, Perona, Ramanan, Doll{\'{a}}r, and Zitnick}]{Lin2014MicrosoftContext}
Lin, T.~Y.; Maire, M.; Belongie, S.; Hays, J.; Perona, P.; Ramanan, D.; Doll{\'{a}}r, P.; and Zitnick, C.~L. 2014.
\newblock {Microsoft COCO: Common objects in context}.
\newblock \emph{ECCV}.

\bibitem[{Liu et~al.(2021)Liu, Huang, Lin, Xie, Wang, Chang, and Liang}]{Liu2021ExploringDistillation}
Liu, L.; Huang, Q.; Lin, S.; Xie, H.; Wang, B.; Chang, X.; and Liang, X. 2021.
\newblock {Exploring Inter-Channel Correlation for Diversity-preserved Knowledge Distillation}.
\newblock \emph{ICCV}.

\bibitem[{Liu et~al.(2019)Liu, Chen, Liu, Qin, Luo, and Wang}]{Liu2019StructuredSegmentation}
Liu, Y.; Chen, K.; Liu, C.; Qin, Z.; Luo, Z.; and Wang, J. 2019.
\newblock {Structured Knowledge Distillation for Semantic Segmentation}.
\newblock \emph{CVPR}.

\bibitem[{Ma, Chen, and Akata(2022)}]{Ma2022DistillingAlignment}
Ma, Y.; Chen, Y.; and Akata, Z. 2022.
\newblock {Distilling Knowledge from Self-Supervised Teacher by Embedding Graph Alignment}.
\newblock \emph{BMVC}.

\bibitem[{Matsubara(2020)}]{Matsubara2020TorchdistillDistillation}
Matsubara, Y. 2020.
\newblock {torchdistill : A Modular, Configuration-Driven Framework for Knowledge Distillation}.

\bibitem[{Miles and Mikolajczyk(2020)}]{Miles2020CascadedSelf-distillation}
Miles, R.; and Mikolajczyk, K. 2020.
\newblock {Cascaded channel pruning using hierarchical self-distillation}.
\newblock \emph{BMVC}.

\bibitem[{Miles, Rodriguez, and Mikolajczyk(2022)}]{Miles2022InformationDistillation}
Miles, R.; Rodriguez, A.~L.; and Mikolajczyk, K. 2022.
\newblock {Information Theoretic Representation Distillation}.
\newblock \emph{BMVC}.

\bibitem[{Miles et~al.(2023)Miles, Yucel, Manganelli, and Saa-Garriga}]{Miles2023MobileVOS:Distillation}
Miles, R.; Yucel, M.~K.; Manganelli, B.; and Saa-Garriga, A. 2023.
\newblock {MobileVOS: Real-Time Video Object Segmentation Contrastive Learning meets Knowledge Distillation}.
\newblock \emph{CVPR}.

\bibitem[{Mobahi, Farajtabar, and Bartlett(2020)}]{mobahi2020selfdistillation}
Mobahi, H.; Farajtabar, M.; and Bartlett, P.~L. 2020.
\newblock Self-Distillation Amplifies Regularization in Hilbert Space.

\bibitem[{Navaneet et~al.(2021)Navaneet, Koohpayegani, Tejankar, and Pirsiavash}]{Navaneet2021SimReg:Distillation}
Navaneet, K.~L.; Koohpayegani, S.~A.; Tejankar, A.; and Pirsiavash, H. 2021.
\newblock {SimReg: Regression as a Simple Yet Effective Tool for Self-supervised Knowledge Distillation}.
\newblock \emph{BMVC}.

\bibitem[{Park et~al.(2019)Park, Corp, Kim, and Lu}]{Park2019RelationalDistillation}
Park, W.; Corp, K.; Kim, D.; and Lu, Y. 2019.
\newblock {Relational Knowledge Distillation}.
\newblock \emph{CVPR}.

\bibitem[{Peng et~al.(2019)Peng, Jin, Li, Zhou, Wu, Liu, Zhang, and Liu}]{Peng2019CorrelationDistillation}
Peng, B.; Jin, X.; Li, D.; Zhou, S.; Wu, Y.; Liu, J.; Zhang, Z.; and Liu, Y. 2019.
\newblock {Correlation congruence for knowledge distillation}.
\newblock \emph{CVPR}.

\bibitem[{Ren et~al.(2022)Ren, Gao, Hua, Xue, Tian, He, and Zhao}]{Ren2022Co-advise:Distillation}
Ren, S.; Gao, Z.; Hua, T.; Xue, Z.; Tian, Y.; He, S.; and Zhao, H. 2022.
\newblock {Co-advise: Cross Inductive Bias Distillation}.
\newblock \emph{CVPR}.

\bibitem[{Romero et~al.(2015)Romero, Ballas, Ebrahimi~Kahou, Chassang, Gatta, and Bengio}]{Romero2015FitNets:Nets}
Romero, A.; Ballas, N.; Ebrahimi~Kahou, S.; Chassang, A.; Gatta, C.; and Bengio, Y. 2015.
\newblock {FitNets: Hints For Thin Deep Nets}.
\newblock \emph{ICLR}.

\bibitem[{Roth et~al.(2021)Roth, Milbich, Ommer, Cohen, and Ghassemi}]{roth2021s2sd}
Roth, K.; Milbich, T.; Ommer, B.; Cohen, J.~P.; and Ghassemi, M. 2021.
\newblock S2SD: Simultaneous Similarity-based Self-Distillation for Deep Metric Learning.

\bibitem[{Russakovsky et~al.(2014)Russakovsky, Deng, Su, Krause, Satheesh, Ma, Huang, Karpathy, Khosla, Bernstein, Berg, and Fei-Fei}]{Russakovsky2014ImageNetChallenge}
Russakovsky, O.; Deng, J.; Su, H.; Krause, J.; Satheesh, S.; Ma, S.; Huang, Z.; Karpathy, A.; Khosla, A.; Bernstein, M.; Berg, A.~C.; and Fei-Fei, L. 2014.
\newblock {ImageNet Large Scale Visual Recognition Challenge}.
\newblock \emph{IJCV}.

\bibitem[{Tian, Krishnan, and Isola(2019)}]{Tian2019ContrastiveDistillation}
Tian, Y.; Krishnan, D.; and Isola, P. 2019.
\newblock {Contrastive representation distillation}.
\newblock \emph{ICLR}.

\bibitem[{Tishby(2015)}]{Tishby2015DeepPrinciple}
Tishby, N. 2015.
\newblock {Deep Learning and the Information Bottleneck Principle}.
\newblock \emph{IEEE Information Theory Workshop (ITW)}.

\bibitem[{Touvron et~al.(2021)Touvron, Cord, Douze, Massa, Sablayrolles, and J{\'{e}}gou}]{Touvron2021TrainingAttention}
Touvron, H.; Cord, M.; Douze, M.; Massa, F.; Sablayrolles, A.; and J{\'{e}}gou, H. 2021.
\newblock {Training data-efficient image transformers {\&} distillation through attention}.
\newblock \emph{PMLR}.

\bibitem[{Tung and Mori(2019)}]{Tung2019Similarity-preservingDistillation}
Tung, F.; and Mori, G. 2019.
\newblock {Similarity-preserving knowledge distillation}.
\newblock \emph{ICCV}.

\bibitem[{Vaswani et~al.(2017)Vaswani, Shazeer, Parmar, Uszkoreit, Jones, Gomez, Kaiser, and Polosukhin}]{Vaswani2017AttentionNeed}
Vaswani, A.; Shazeer, N.; Parmar, N.; Uszkoreit, J.; Jones, L.; Gomez, A.~N.; Kaiser, L.; and Polosukhin, I. 2017.
\newblock {Attention Is All You Need}.
\newblock \emph{NeurIPS}.

\bibitem[{Wang et~al.(2019)Wang, Yuan, Zhang, and Feng}]{Wang2019DistillingImitationb}
Wang, T.; Yuan, L.; Zhang, X.; and Feng, J. 2019.
\newblock {Distilling Object Detectors with Fine-grained Feature Imitation}.
\newblock \emph{CVPR}.

\bibitem[{Xu et~al.(2020)Xu, Liu, Li, and Loy}]{Xu2020KnowledgeSelf-supervision}
Xu, G.; Liu, Z.; Li, X.; and Loy, C.~C. 2020.
\newblock {Knowledge Distillation Meets Self-supervision}.
\newblock \emph{ECCV}.

\bibitem[{Yang et~al.(2021)Yang, An, Cai, and Xu}]{Yang2021HierarchicalDistillation}
Yang, C.; An, Z.; Cai, L.; and Xu, Y. 2021.
\newblock {Hierarchical Self-supervised Augmented Knowledge Distillation}.
\newblock \emph{IJCAI}.

\bibitem[{Yim(2017)}]{Yim2017ALearning}
Yim, J. 2017.
\newblock {A Gift from Knowledge Distillation: Fast Optimization, Network Minimization and Transfer Learning}.
\newblock \emph{CVPR}.

\bibitem[{Zagoruyko and Komodakis(2019)}]{Zagoruyko2019PayingTransfer}
Zagoruyko, S.; and Komodakis, N. 2019.
\newblock {Paying more attention to attention: Improving the performance of convolutional neural networks via attention transfer}.
\newblock In \emph{ICLR}.

\bibitem[{Zbontar et~al.(2021)Zbontar, Jing, Misra, LeCun, and Deny}]{Zbontar2021BarlowReduction}
Zbontar, J.; Jing, L.; Misra, I.; LeCun, Y.; and Deny, S. 2021.
\newblock {Barlow Twins: Self-Supervised Learning via Redundancy Reduction}.
\newblock \emph{ICML}.

\bibitem[{Zhang et~al.(2019)Zhang, Song, Gao, Chen, Bao, and Ma}]{Zhang2019BeDistillation}
Zhang, L.; Song, J.; Gao, A.; Chen, J.; Bao, C.; and Ma, K. 2019.
\newblock {Be Your Own Teacher: Improve the Performance of Convolutional Neural Networks via Self Distillation}.

\bibitem[{Zhang, Isola, and Efros(2016)}]{Zhang2016ColorfulColorization}
Zhang, R.; Isola, P.; and Efros, A.~A. 2016.
\newblock {Colorful Image Colorization}.

\bibitem[{Zhang et~al.(2020)Zhang, Lan, Dai, Zeng, and Bai}]{Zhang2020Prime-AwareDistillation}
Zhang, Y.; Lan, Z.; Dai, Y.; Zeng, F.; and Bai, Y. 2020.
\newblock {Prime-Aware Adaptive Distillation}.
\newblock \emph{ECCV}.

\bibitem[{Zhang and Sabuncu(2020)}]{zhang2020selfdistillation}
Zhang, Z.; and Sabuncu, M.~R. 2020.
\newblock Self-Distillation as Instance-Specific Label Smoothing.

\bibitem[{Zhao, Song, and Qiu(2022)}]{Zhao2022DecoupledDistillation}
Zhao, B.; Song, R.; and Qiu, Y. 2022.
\newblock {Decoupled Knowledge Distillation}.
\newblock \emph{CVPR}.

\bibitem[{Zhu et~al.(2021{\natexlab{a}})Zhu, Tang, Chen, and Yu}]{Zhu2021ComplementaryDistillation}
Zhu, J.; Tang, S.; Chen, D.; and Yu, S. 2021{\natexlab{a}}.
\newblock {Complementary Relation Contrastive Distillation}.
\newblock \emph{CVPR}.

\bibitem[{Zhu et~al.(2021{\natexlab{b}})Zhu, Lyu, Wang, and Zhao}]{Zhu2021TPH-YOLOv5:Scenarios}
Zhu, X.; Lyu, S.; Wang, X.; and Zhao, Q. 2021{\natexlab{b}}.
\newblock {TPH-YOLOv5: Improved YOLOv5 Based on Transformer Prediction Head for Object Detection on Drone-captured Scenarios}.
\newblock \emph{VisDrone ICCV workshop}.

\end{thebibliography}
